\pdfoutput=1

\documentclass[11pt]{article}

\usepackage[final]{ACL2023}

\usepackage{times}
\usepackage{latexsym}
\usepackage{graphicx}
\usepackage[T1]{fontenc}

\usepackage[utf8]{inputenc}

\usepackage{microtype}

\usepackage{inconsolata}
\usepackage{multirow}
\usepackage{booktabs}

\title{i-Code Studio: A Configurable and Composable Framework \\for Integrative AI}

\author{Yuwei Fang\thanks{Co-first authors.}, Mahmoud Khademi\footnotemark[1], Chenguang Zhu, Ziyi Yang, Reid Pryzant, Yichong Xu, \\ \bf Yao Qian, Takuya Yoshioka, Lu Yuan, Michael Zeng and Xuedong Huang \\
Microsoft Cognitive Services Research Group \\
\small \{\tt yuwfan, mkhademi, chezhu\}@microsoft.com}

\begin{document}
\maketitle
\begin{abstract}
Artificial General Intelligence (AGI) requires comprehensive understanding and generation capabilities for a variety of tasks spanning different modalities and functionalities. Integrative AI is one important direction to approach AGI, through combining multiple models to tackle complex multimodal tasks. However, there is a lack of a flexible and composable platform to facilitate efficient and effective model composition and coordination. In this paper, we propose the i-Code Studio, a configurable and composable framework for Integrative AI. The i-Code Studio orchestrates multiple pre-trained models in a finetuning-free fashion to conduct complex multimodal tasks. Instead of simple model composition, the i-Code Studio provides an integrative, flexible, and composable setting for developers to quickly and easily compose cutting-edge services and technologies tailored to their specific requirements. The i-Code Studio achieves impressive results on a variety of zero-shot multimodal tasks, such as video-to-text retrieval, speech-to-speech translation, and visual question answering. We also demonstrate how to quickly build a multimodal agent based on the i-Code Studio that can communicate and personalize for users. The project page with demonstrations and code is at \url{https://i-code-studio.github.io/}
\end{abstract}

\section{Introduction}
Large language models (LLMs) such as BERT~\cite{devlin2018bert} and GPT-3~\cite{brown2020language}, visual-language models (VLMs) like CLIP~\cite{radford2021learning} and DALL-E~\cite{ramesh2021zero}, and audio language models (ALMs) such as W2V-BERT~\cite{chung2021w2v} have enabled a variety of capabilities, from zero-shot image classification to reading comprehension, automatic speech recognition, and photorealistic image generation. The performance and capability of these pre-trained models are, however, influenced by the data they are exposed to, which varies across different domains; LLMs are trained on diverse sources of data, such as webpages, novels, and Wikipedia corpora, while VLMs are trained on pairs of images or videos and their captions, and ALMs are trained on audio data such as speech. 
\begin{figure}[h]
\centering
{\includegraphics[width=0.98\linewidth]{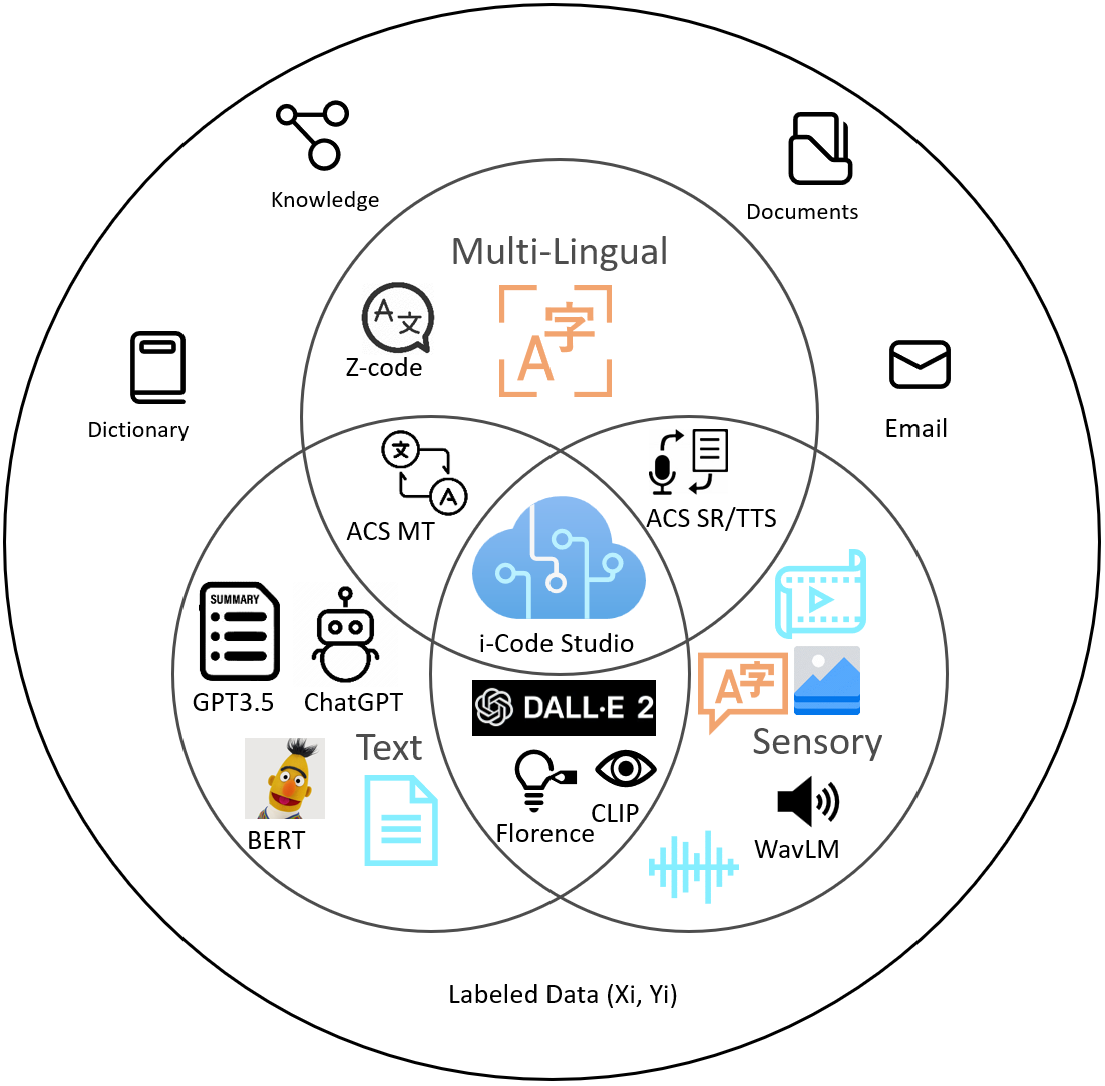}}
\caption{\label{fig:framework} The i-code Studio is a configurable and composable architecture for integrative AI allowing developers to quickly and easily orchestrate various cutting-edge pre-trained models in a finetuning-free fashion.
}
\end{figure}
These distinct training domains render the pre-trained models different and sometimes complementary capabilities. For instance, LLMs are suitable for tasks such as reading comprehension but unable to interpret audio and visual information; VLMs can produce photorealistic images but cannot tackle complex language understanding. On the other hand, humans can often easily handle distinct tasks like the above with multimodal input and output. Therefore, in order to build Artificial General Intelligence (AGI), we need to break the barriers between modalities and specific tasks. 

Instead of building a single model to handle all possible tasks, which is infeasible under current technology, a lot of research has recently emerged to focus on the composition of large pre-trained models to achieve integrative AI, either via fine-tuning them jointly on new tasks~\cite{yang2022code, hu2021unit, wang2021simvlm, alayrac2022flamingo, tewel2022zerocap}, or via a shared modality, such as language, to capture new multimodal capabilities without the need for finetuning~\cite{tewel2022zerocap, zeng2022socratic, wang2022language, li2022composing}. Issues with these approaches are that 1) there often lacks data and computation resources for joint finetuning, and 2) one cannot easily configure and compose different large pre-trained models in an agile framework to adapt to different needs. Therefore, in this paper, we propose the i-Code Studio, a configurable and composable framework for integrative AI (Figure~\ref{fig:framework}) . 
The i-Code Studio allows developers to quickly and easily orchestrate various cutting-edge pre-trained models in a finetuning-free fashion.

These pre-trained models are from different modalities, and the strength of each individual model is integrated to conduct complex multimodal tasks. For each task, a directed acyclic graph (DAG) is configured so that the related models cooperate to produce the desired output. The input data flows through each node in the DAG, enabling complex multimodal tasks to be completed. This makes i-Code Studio an integrative, flexible, and composable framework. For instance, for visual question answering task, a DAG is configured using the input image, the input question, the Florence~\cite{yuan2021florence} vision foundation model, a language prompt, the ChatGPT, and an output, each represented by a node. The visual information from the input image is fed into Florence. The Florence node processes the image and outputs a set of detected object categories/tags and a caption. These outputs and the input question are then fed into a node that generates a VLM-informed language prompt. Finally, this cross-modal prompt is used by ChatGPT to generate an answer to the input question which is sent to the output node.

In this paper, we showcase the effectiveness of the i-Code Studio using models from Azure Cognitive Services (ACS) and OpenAI services. The resulting integrative model achieves the state-of-the-art (SOTA) or comparable to the SOTA performance on zero-shot tasks such as speech-to-speech translation, video-to-text retrieval, and visual question answering. We also show how to quickly build a multimodal agent to interact with a user. In summary, our main contributions are the following:

(1) We propose i-Code Studio, a new integrative, configurable, and composable framework which can be used to compose various pre-trained models.

(2) We show how i-Code Studio can achieve impressive results on a variety of zero-shot multimodal tasks, e.g. video-to-text retrieval, speech-to-speech translation, and visual question answering.

(3) We utilize i-Code Studio to build a multimodal agent that can communicate and personalize for users by leveraging ACS and OpenAI services. 

\section{Related Work}
 Recently, the composition of large pre-trained models has been extensively studied. The most common way to compose these models is to fine-tune them jointly on new tasks.~\citet{hu2021unit} proposed UniT, a Unified Transformer model that is capable of learning several tasks across multiple domains, including object detection and multimodal reasoning. This model is based on a transformer encoder-decoder architecture, where each input modality is encoded with an encoder, and shared decoders are used to make predictions for each task.~\citet{wang2021simvlm} proposed a Vision-Language Pretraining framework, called SimVLM that is trained end-to-end with a single language modeling objective. The SimVLM reduces the complexity of training by utilizing weak supervision on a large scale.~\citet{alayrac2022flamingo} proposed Flamingo, a collection of VLMs that can connect pre-trained vision-only and language-only models, process sequences of interleaved visual and textual data, and accept images or videos as inputs. However, these methods can be computationally expensive. The i-Code Studio differs from these approaches since it does not require finetuning, which enables the fast composition of pre-trained models for a variety of tasks and reduces the time and expense associated with finetuning. 
 
Unlike these work, models can be composed via a shared modality, such as language.~\citet{tewel2022zerocap} combined a visual-semantic model with a large language model, enabling the models to take advantage of the knowledge present in both web-scale models for image caption generation task. More related to our work,~\citet{zeng2022socratic} proposed Socratic Models, a modular framework that enables multiple pre-trained models to exchange information with each other, capture new multimodal capabilities without the need for finetuning, and be composed without any prior training using multimodal-informed prompting. Our i-Code Studio is a more integrative, flexible, and composable framework compared to these work, allowing users to compose cutting-edge models and technologies customized for their particular needs easily.

Distinct from the work mentioned,~\citet{li2022composing} proposed a closed-loop approach to combining pre-trained models in such a way that they act as generators and scorers. The generators create proposals, while the scorers provide feedback to improve the generated results. This type of iterative consensus optimization allows models to correct mistakes made by other models, leading to significant improvements in downstream tasks.~\cite{huang2022inner} studied the application of LLMs in embodied environments for robotic control. They combined LLMs with different sources of text feedback and found that natural language acts as a universal means of communicating with the model. The resulting system, called Inner Monologue, integrates various components such as perception models, robotic skills, and human feedback to effectively execute user commands. 

\section{The i-Code Studio Framework}

\begin{figure}[t!]
\centering
{\includegraphics[width=0.98\linewidth]{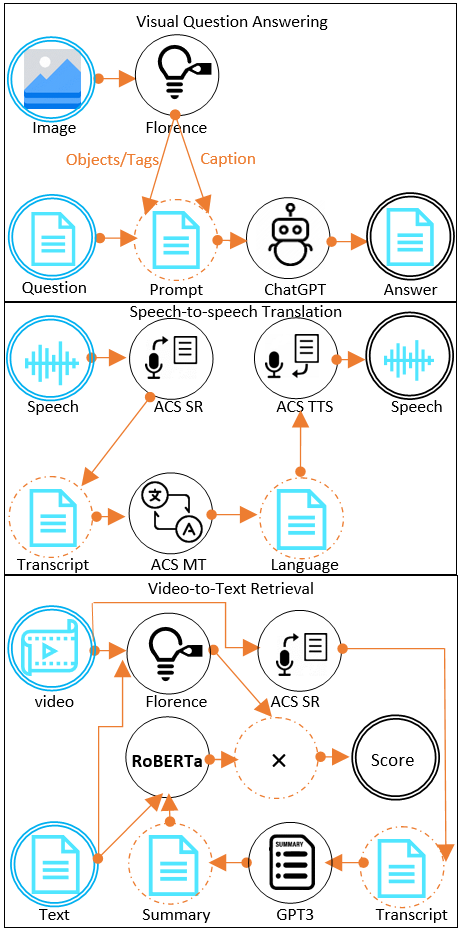}}
\caption{\label{fig:i-code} The i-Code Studio can be used to build AI solutions for various multimodal tasks. For each task, a DAG is configured so that the related models cooperate to produce the desired output. The input data flows through each node in the DAG, enabling complex multimodal tasks to be completed. The input nodes are represented by double blue circles, the model/service nodes, e.g. ChatGPT and Florence, by black circles, the output nodes by double black circles, and the rest by dash-dotted red circles. See the text for details about each multimodal task.}
\end{figure}
In this section, we introduce i-Code Studio, a configurable and composable framework for integrative AI. Given a complex multimodal task, the i-Code Studio provides a generic framework for developers to quickly and easily integrate and compose several large pre-trained models and services across different modalities without any training or finetuning to accomplish the task. Figure~\ref{fig:i-code} shows examples of building AI solutions for various multimodal tasks using the i-Code Studio framework. For each task, the framework can be represented via a DAG, where the nodes with no incoming edge are the raw input data such as image, text, video and speech, the nodes with no outgoing edges are the outputs of the given task, and the rest of the nodes are foundation models/services or hold intermediate model outputs from other models/services. The input to a node comes from the raw input, and/or the output from previous nodes. The input data flows through each node in the DAG, enabling complex multimodal tasks to be completed. An outgoing edge from a model/service node represent an API provided by the model/service. For each task, the inputs enter the DAG from the input nodes, and are processed by one or more models or model services. In the process, edges convert the format of a module's output, filters data, or apply an API such as summarization, translation, object detection, image captioning, transcribing, text-to-speech synthesis, etc. 

For each task, a DAG is configured so that the related models cooperate to produce the desired output. The different components of i-Code Studio cooperate seamlessly to form a single, integrated AI solution, and can be adjusted to fit the specific needs of the user. For instance, for visual questions answering (VQA) task the input is an image and a question related to the image (see Figure~\ref{fig:i-code}).
We can first apply image captioning and object detection services to the input image. The output text, which contains the visual information, is merged with the input question as the prompt to ChatGPT, which answers the question. For speech-to-speech translation, the DAG is configured with Speech Recognition (SR) $\rightarrow$ Machine Translation (MT) $\rightarrow$ Text-To-Speech (TTS). This DAG transcribes the source speech, translates the transcription into the target language, and generates the target speech.

To build i-Code Studio, we utilize Azure Machine Learning Studio, a cloud-based, collaborative, drag-and-drop development environment for building, testing, and deploying machine learning models. We encapsulate available models and services from Azure Cognitive Services (ACS) as independent APIs and deploy them as an integrated web service for real-time invoking. In this way, it allows developers to flexibly combine them to build their own applications. More details about the available foundation models and services are presented in Appendices ~\ref{sec:appendix-a} and~\ref{sec:appendix-b}.
 

\section{Evaluations}
In this section, we presents our experiments in three tasks covering language, speech and vision modality: 1) video-to-text retrieval; 2) visual question answering and 3) speech-to-speech translation.
\subsection{Video-to-Text Retrieval}
Video-to-Text retrieval task is to select the most relevant text from a pool of candidates given the video, which typically involves all modalities across language, vision and speech. Thus, it can be an ideal task to test the capabilities of i-Code Studio. Following ~\citet{zeng2022socratic}, the pipeline is organized into the following steps: ($i$) calculate the similarity score $s_1$ between the average vision features of video and text features of captions via ACS Vision service~\cite{yuan2021florence}; 
($ii$) calling ACS Speech service to transcribe the video to text; ($iii$) summarize the transcript with Azure OpenAI services using GPT-3~\cite{brown2020language}; ($iv$) compute a text-based similarity score $s_2$ between the generated summary and the captions with pre-trained language model; ($v$) compute the final relevance score $s=s_1\times s_2$, combining vision-text based score and speech-text based score; ($vi$) select the text with the highest relevance score as answer.

\begin{table}[t!]
\small
\centering
\setlength\tabcolsep{1.5pt}
\resizebox{1.0\columnwidth}{!}{
\begin{tabular}{llcccc}
\toprule
 & Method & R@1 & R@5 & R@10 \\
 \midrule
\multirow{3}{*}{Finetuned} & JMEC~\cite{jmec} & 12.5 & 32.1 & 42.4 \\
& Collab. Experts\cite{Liu2019a} & 15.6 & 40.9 & 55.2 \\
& CLIP2Video~\cite{fang2021clip2video} & \textbf{54.6} & \textbf{82.1} & \textbf{90.8} \\ \midrule
\multirow{3}{*}{Zero-shot} & CLIP \cite{clip_via} & 40.3 & 69.7 & 79.2 \\ 
& SMs~\cite{zeng2022socratic} & 44.7 & 71.2 & 80.0 \\ 
& i-Code Studio & \textbf{49.8} & \textbf{74.8} & \textbf{82.2} \\ \bottomrule
\end{tabular}
}
\caption{Video-to-text retrieval results on MSR-VTT~\cite{xu2016msr} dataset.}
\label{table:msrvtt}
\end{table}

Table~\ref{table:msrvtt} shows our results on MSR-VTT~\cite{xu2016msr}, which is the most popular large-scale dataset for video-to-text retrieval and consists of 10,000 video clips from 20 categories, and each video clip is annotated with 20 English sentences by Amazon Mechanical Turks. We use the standard recall metrics for evaluation and compare our approach with both finetuned and zero-shot methods. We can see that in zero-shot setting, i-Code Studio outperforms previous state-of-the-art (SOTA) SMs by 5.1 points in R@1, thus achieving the new SOTA in this setting. Compared with finetuned approach, i-Code Studio significantly narrowed the gap between the zero-shot and fine-tuned approach, showing the promising of the zero-shot approach.

\subsection{Visual Question Answering}
\begin{figure}[ht]
\centering
{\includegraphics[width=0.96\linewidth]{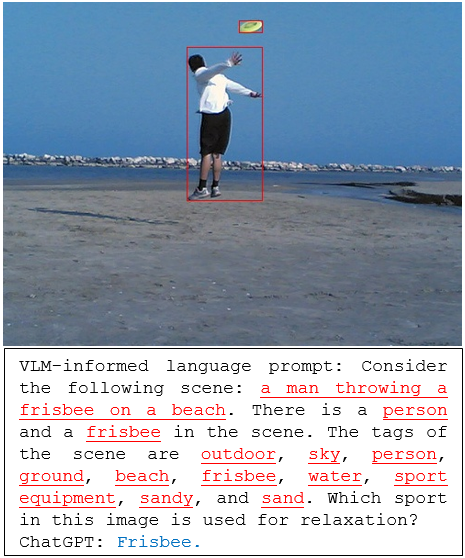}}
\caption{\label{fig:vqa} VQA with i-Code Studio: a VLM-informed language prompt is created using Florence outputs and input question. The red underlined text are the caption, object categories, and tags detected by Florence. The prompt is then fed into ChatGPT to predict an answer. 
}
\end{figure}

The i-Code Studio can be used to answer visual questions (see Figure~\ref{fig:vqa}). Specifically, Azure Cognitive Services' Florence~\cite{yuan2021florence} is used to zero-shot detect a set of object categories in the input image, generate a set of tags associated to it, and create a caption that describes the image. These descriptions and the input question are then used to form a VLM-informed language prompt, which is fed into ChatGPT to predict an answer. We evaluated i-Code Studio's performance on the FVQA dataset~\cite{wang2017fvqa} for the visual question answering task. FVQA is a VQA dataset that mostly contains questions requiring external knowledge to answer, and provides supporting fact triplets alongside the image-question-answer triplets. Following~\cite{wang2017fvqa}, we used $1,090$ test images, amounting to $2,899$ questions. Our results are presented in Table~\ref{table:vqa}. The i-Code Studio significantly outperforms Fact-based VQA without the support facts from the dataset, likely due to the power of Florence's vision foundation model and ChatGPT's capability to answer questions requiring external knowledge.
\begin{table}[htbp]
\small
\centering
\setlength\tabcolsep{1.5pt}
\begin{tabular}{lc}
\toprule
 Method & Accuracy \\
 \midrule
Human & 77.99 \\ 
\midrule
Fact-based VQA~\cite{wang2017fvqa} & 56.91 \\
Fact-based VQA (Ensemble)~\cite{wang2017fvqa} & 58.76 \\
i-Code Studio & \textbf{60.59} \\ \bottomrule
\end{tabular}
\caption{VQA results on FVQA dataset.}
\label{table:vqa}
\end{table}

\subsection{Speech-to-speech Translation}
Speech-to-speech translation task consists of transcribing spoken language into text, translating the text into another language, and then generating speech in the target language. We use this task to evaluate the multilingual and speech capabilities of i-Code Studio. Specifically, we first leverage ACS Speech Recognition service to transcribe the incoming speech, then use ACS Language Machine Translation service to translate in the target languages, and finally call ACS Text-To-Speech to synthesize the speech in the target languages.

We evaluate i-Code Studio on CVSS~\cite{jia2022cvss} dataset, a massively multilingual-to-English speech-to-speech translation corpus. It covers sentence-level parallel speech-to-speech translation pairs from 21 languages into English and is derived from the Common Voice speech corpus~\cite{ardila-etal-2020-common} and the CoVoST 2~\cite{wang2020covost} speech-to-text translation corpus. The translation speech in CVSS is synthesized with two state-of-the-art TTS models trained on the LibriTTS corpus. As the speech generation quality is measured by human in mean opinion score (MOS) on naturalness and speaker similarity metrics, here we only report translated text result in BLEU metric using SacreBLEU with its default configuration. Following~\citet{jia2022cvss}, we group the evaluation results on high-resource source languages (French, German, Catalan and Spanish) and low-resource ones (all the rest). From Table~\ref{table:cvss}, we can see the i-Code Studio outperforms previous SOTAs significantly by 22.5 points on average. The improvement of high-resource languages still has about 8.3 points, demonstrating the strong capabilities of the i-Code Studio framework.
\begin{figure*}[htbp]
\centering
{\includegraphics[width=0.9\linewidth]{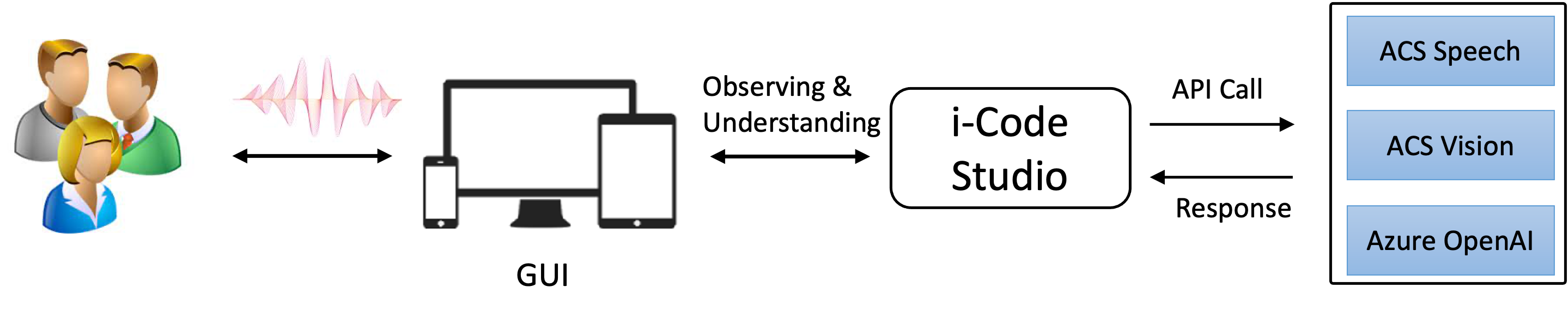}}
\caption{\label{fig:agent} An overview of the multimodal agent which is built using the i-Code Studio.
}
\end{figure*}

\begin{table}[t!]
\small
\centering
\setlength\tabcolsep{1.5pt}
\begin{tabular}{l|ccc}
\toprule
Model & All & Hi-Res & Lo-Res  \\
\midrule
\citet{li-etal-2021-multilingual} (Scratch-BL) & - & 14.8 & –  \\
\citet{wang2021covost} (A2A-L) &  7.5 & 24.0 & 3.7 \\
\citet{wang2021covost} (A2E-M, arXiv) & - & 24.5 & - \\
~\citet{jia2022cvss} & 11.0 & 29.4 & 6.7 \\ 
~\citet{jia2022cvss} (ASR pre-training) & 13.3 & 31.4 & 9.0 \\
\midrule
i-Code Studio & 35.8 & 39.7 & 34.8 \\
\bottomrule
\end{tabular}
\caption{Speech-to-text evaluation results on CVSS dataset. We call ACS Speech Recognition, ACS Machine Translation, and ACS Text-to-Speech services in a cascade approach. Hi-Res and Lo-Res stand for high-resource and low-resource languages respectively.}
\label{table:cvss}
\end{table}

\section{Applications: Multimodal Agents}

\begin{figure*}[t!]
\centering
{\includegraphics[width=0.99\linewidth]{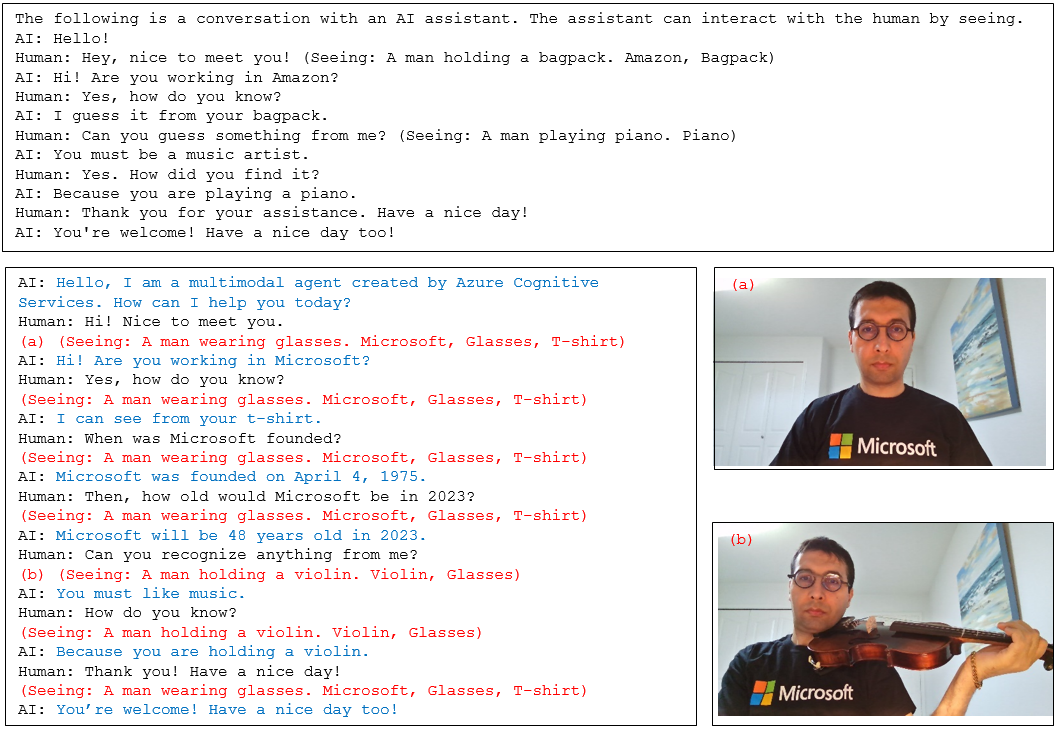}}
\caption{\label{fig:demo} The i-Code Studio can be used to build a multimodal virtual assistant. During the conversation the user input and history context are prepended with the captions/tags from Florence vision (shown in red) and fed as input into GPT-3. The bottom boxes show the conversation as well as two snapshots of the input video from the camera. 
}
\end{figure*}

As humans, we have a complex sensory system that allows us to experience the world around us. We use our eyes to see, ears to hear,  mouths to talk, and brains to process and interpret the information we receive. Inspired by this, we utilize i-Code Studio to build a multimodal agent that can communicate and personalize for users. Specifically, the eyes of the agent use Azure Vision services to interpret visual images signals and send signals to the brain; the ears and mouth use Azure Speech services to collect sound waves and produce sounds; the brain leverage Azure OpenAI services to integrate all the sensory signals received from the eyes, ears and uses them to make decisions. This interconnected system of sensory organs and the brain are what enables our multimodal agents to understand and interact with the world around us. Our multimodal agent is a virtual assistant with ``eyes'' (Florence), ``ears'' (ACS ASR), ``brain'' (e.g. ChatGPT and GPT-3) and mouth (ACS TTS). The i-Code Studio integrates speech and vision signals from users by composing and configuring services from ACS and OpenAI. Figure~\ref{fig:demo} shows a demo example. Using VLM-infromed language prompting, i-Code Studio can enable multimodal dialogue between the user and agent. GUIs call i-Code Studio once to simplify the developing cost while giving consistent user experience.



\section{Conclusion}
The i-Code Studio, is a new configurable and composable framework for Integrative AI. It orchestrates multiple pre-trained models to conduct complex multimodal tasks, without the need for finetuning. We showed the i-Code Studio can achieve impressive results on three multimodal tasks. We also demonstrated how to build a multimodal virtual assistant agent with the i-Code Studio. With further research and development, the i-Code Studio can be extended to be more flexible and powerful to create even more complex applications.
\section{Screencast Video}
In this section, the public link to one of our example demos for the multimodal agent is provided\footnote{\url{https://drive.google.com/file/d/10lEZQ9LbQpR_kc8zsmenRXjCQejSO-G3/view?usp=share_link}}.
\section{Limitations}
The i-Code Studio currently relies on a limited number of pre-trained models and services. While this is sufficient for many multimodal tasks, the framework needs additional services to support more complex multimodal tasks. Moreover, to demonstrate the capabilities of the i-Code Studio, we need to apply the framework to more complex multimodal tasks such as meeting summarization and image generation from textual descriptions. 

\bibliography{anthology,custom}
\bibliographystyle{acl_natbib}

\clearpage
\appendix

\section{Foundation Models}
\label{sec:appendix-a}
Foundation models, first introduced by ~\citet{bommasani2021opportunities}, 
refer to any model that is pre-trained on broad data at scale and can be adapted to a wide range of downstream tasks. 
As a general paradigm of AI, foundation models have shown impressive performances and generalization capabilities in various modalities~\cite{brown2020language, clip, yuan2021florence}.
\paragraph{Large Language Models}
Large language models (LMs), trained on massive text collections such as BERT~\cite{devlin-etal-2019-bert}, GPT-2~\cite{radford2019language}, DeBERTa~\cite{he2021deberta}, achieve state-of-the-art performances on many natural language processing benchmarks. More recent works, like GPT-3~\cite{brown2020language}, OPT~\cite{zhang2022opt}, PaLM~\cite{chowdhery2022palm}, Chinchilla~\cite{hoffmann2022training}, have shown surprising emergent capabilities to generate text and can be ``prompted'' to perform a range of language tasks given zero or few examples of the task as input. In the i-Code Studio framework, we include three language-based foundation models to support diverse tasks and applications: Z-Code~\footnote{\url{https://www.microsoft.com/en-us/research/project/project-zcode/}} for multilingual tasks like machine translation, GPT-3~\cite{brown2020language} and ChatGPT~\footnote{\url{https://chat.openai.com/}} for general NLP tasks like text summarization and question answering. 

\paragraph{Vision Language Models}
Vision language models (Vision LMs), trained on web-scale image-text and video data, such as CLIP~\cite{clip}, ALIGN~\cite{jia2021scaling}, DALL-E~\cite{ramesh2021zero}, Imagen~\cite{saharia2022photorealistic} and Nuwa-infinity~\cite{liang2022nuwainfinity}, demonstrate superior performance on various computer vision tasks, such as classification, retrieval, object detection, VQA, image caption, video retrieval and action recognition. In Azure Cognitive Services, the Project Florence~\footnote{\url{https://www.microsoft.com/en-us/research/project/projectflorence/}} is initiated to advance state-of-the-art computer vision technologies and develop the next-generation framework for visual recognition. Specifically, Florence~\cite{yuan2021florence} is trained on noisy Web-scale data end-to-end with a unifying objective, allowing the model to achieve state-of-the-art performances across a wide range of benchmarks. In i-Code Studio, Florence is utilized as the vision foundation model.

\paragraph{Audio Language Models} 
Audio language models leverage discretized audio tokens/codes to train a model by using a language modeling task, such as w2v-BERT~\cite{chung2021w2v}, WavLM~\cite{9814838}, and Vall-E~\cite{wang2023neural}, and bring significant improvements for various speech processing tasks like speech-to-text, text-to-speech, speaker recognition/diarization, speech separation, etc. In Azure Cognitive Speech Services, speech models were trained by using more than a few hundred of thousand hours of speech audio in a manner of supervised learning. 

\section{Machine Learning Services}
\label{sec:appendix-b}
A machine learning service is usually built on top of the foundation models, provide a comprehensive suite of cloud-based artificial intelligence (AI) and machine learning (ML) tools and services. These tools provide developers with easy-to-use, pre-built algorithms and APIs that can be integrated into a wide range of applications. The i-Code Studio adopt Azure Cognitive Services\footnote{\url{https://azure.microsoft.com/en-us/products/cognitive-services/\#overview}}, which provides a variety of models and services for different modalities.
Developers can easily leverage Azure Cognitive services to add intelligence features to their applications, such as sentiment analysis, object detection, speech recognition and text-to-speech, without having to build the AI models from scratch

We include the following services for each modality in one framework so that our architecture can flexibly enable complicated applications that are difficult to create with an end-to-end approach and meanwhile provide users with a consistent experience.
The i-Code Studio adopts the design of prompt learning [cite] to quickly adapt the architecture to different tasks through informed multimodal prompting with just a few labeled examples.

\paragraph{Language}
Azure Cognitive Services for Language (ACS Language) is a cloud-based service that provides Natural Language Processing (NLP) features for understanding and generation by using REST APIs and client libraries. Using Z-Code as the backbone,  the language services provide the following functionalities: natural language understanding, question answering, text summarization and machine translation. Besides, we also integrate Azure OpenAI Services which use ChatGPT, GPT-3, Codex and Embeddings from OpenAI as the backbone to enable new reasoning and comprehension capabilities for building cutting-edge applications. Specifically, in our architecture, we include three language APIs: ($i$) machine translation: translating text from one language to another. This can be used to realize multilingual communication between human and machines. ($ii$) ChatGPT: an interactive dialogue language model; ($iii$) GPT-3: capable of a wide range of NLP tasks such as text generation, translation, summarization and question answering.\footnote{For GPT-3, We use text-davinci-003 model for downstream tasks and applications.}
\paragraph{Speech}

Azure Cognitive Speech Service (ACS Speech) provides speech capabilities with an Azure Speech resource. It can accurately transcribe multi-lingual speech-to-text, produce text-to-speech with real human-like voices, translate spoken audio, and correctly identify the speakers in conversations. We integrate two speech APIs in our architecture: $(i)$ Speech-to-Text, to transcribe your speech to text in real-time or to transcribe recorded audio files to text; $(ii)$ Text-to-Speech, to convert input text into synthetic speech in real-time  or to generate audio files from text with either prebuilt or customized natural voice. 

\paragraph{Vision}
Azure Cognitive Services for Vision (ACS Vision) are a set of services offered by Microsoft Azure that allow developers to add computer vision capabilities to their applications. It provides a range of services for tasks such as object detection and recognition, image analysis, optical character recognition (OCR), and facial recognition. We integrate two vision APIs in our architecture: $(i)$ object detection: identify objects in an image and locate the bounding box  within the frame. $(ii)$ image captioning: generate a description of an entire image in human-readable language, using complete sentences. 

\end{document}